\documentclass[runningheads]{llncs}
\usepackage{graphicx}

\usepackage{tikz}
\usepackage{comment}
\usepackage{amsmath,amssymb} 
\usepackage{color}
\usepackage{subfigure}
\usepackage{multirow}
\usepackage[misc]{ifsym}
\usepackage[colorlinks,
            linkcolor=red,
            anchorcolor=green,
            citecolor=green]{hyperref}

\usepackage[accsupp]{axessibility}  

\begin{document}
\pagestyle{headings}
\mainmatter
\def\ECCVSubNumber{3452}  

\title{Overlooked Poses Actually Make Sense: Distilling Privileged Knowledge for Human Motion Prediction} 

\titlerunning{Distilling Privileged Knowledge for Human Motion Prediction}
%
\author{Xiaoning Sun\inst{1} \and
Qiongjie Cui\inst{1} \and
Huaijiang Sun\inst{1}\textsuperscript{\Letter} \and
Bin Li\inst{2} \and
Weiqing Li\inst{1} \and
Jianfeng Lu\inst{1}}
\authorrunning{X. Sun et al.}
%
\institute{Nanjing University of Science and Technology, China
\email{\{sunxiaoning,cuiqiongjie,sunhuaijiang,li\_weiqing,lujf\}@njust.edu.cn} \and
Tianjin AiForward Science and Technology Co., Ltd., China
\email{libin@aiforward.com}}

\maketitle

\begin{abstract}
  Previous works on human motion prediction follow the pattern of building a mapping relation between the sequence observed and the one to be predicted. However, due to the inherent complexity of multivariate time series data, it still remains a challenge to find the extrapolation relation between motion sequences. In this paper, we present a new prediction pattern, which introduces previously overlooked human poses, to implement the prediction task from the view of interpolation. These poses exist \emph{after} the predicted sequence, and form the privileged sequence. To be specific, we first propose an InTerPolation learning Network (ITP-Network) that encodes both the observed sequence and the privileged sequence to interpolate the in-between predicted sequence, wherein the embedded Privileged-sequence-Encoder (Priv-Encoder) learns the privileged knowledge (PK) simultaneously. Then, we propose a Final Prediction Network (FP-Network) for which the privileged sequence is not observable, but is equipped with a novel PK-Simulator that distills PK learned from the previous network. This simulator takes as input the observed sequence, but approximates the behavior of Priv-Encoder, enabling FP-Network to imitate the interpolation process. Extensive experimental results demonstrate that our prediction pattern achieves state-of-the-art performance on benchmarked H3.6M, CMU-Mocap and 3DPW datasets in both short-term and long-term predictions.
\end{abstract}

\section{Introduction}

Human motion prediction, aimed at forecasting future poses according to the historical motion sequence, has been widely applied in the field of autonomous driving \cite{1}, human-machine interaction \cite{2} and action detection \cite{3}. To make the best of human body structure information, recent methods employ graph convolutional networks (GCNs) \cite{27,29,30-2,60} to model spatial dependencies or topological relations of body joints.

\begin{figure}[t]
  \centering
  \includegraphics[width=0.96\linewidth]{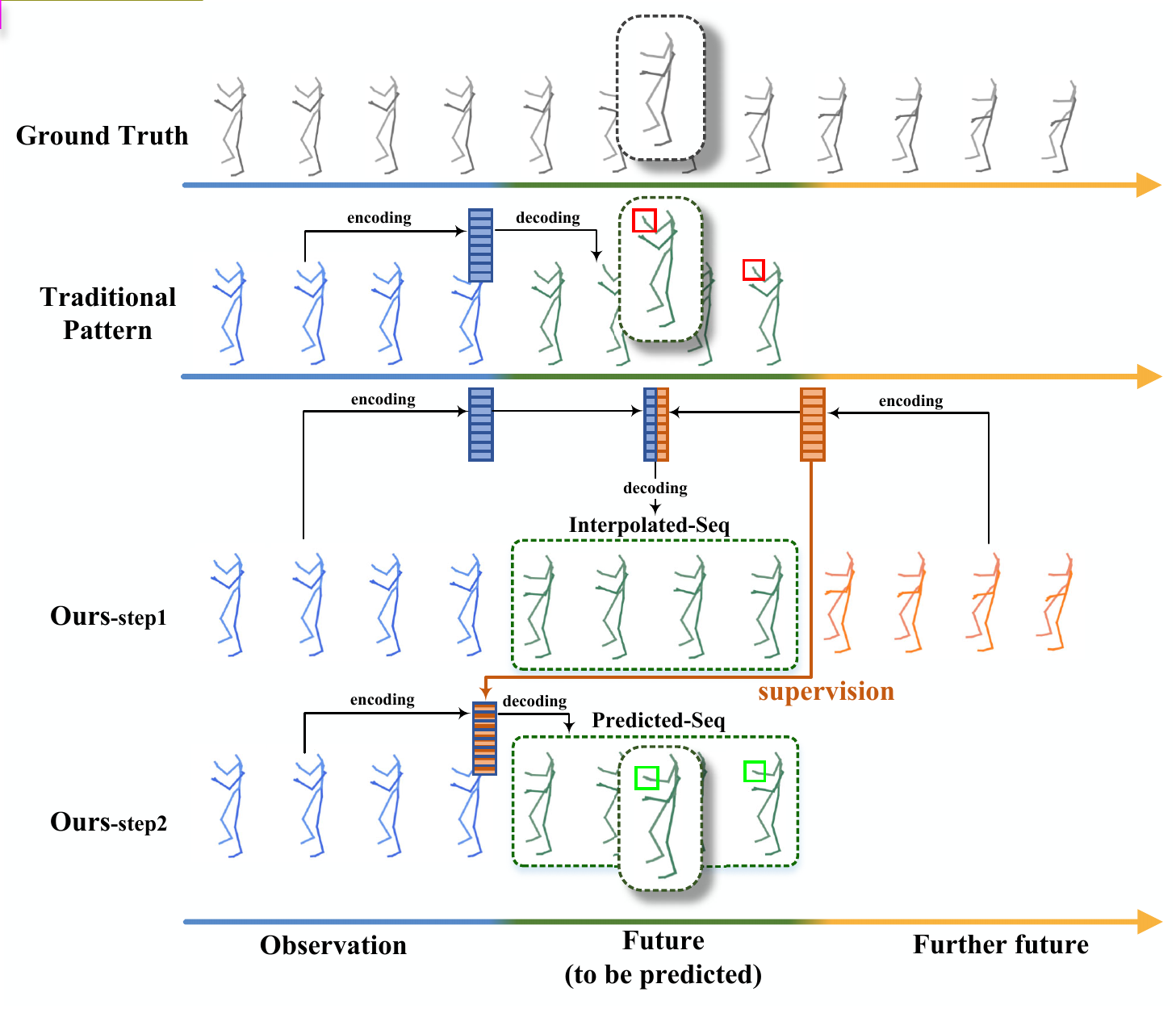}

   \caption{Comparison between the traditional human motion prediction pattern and ours. Previous works predict future motions on historical ones by direct extrapolation, which may bring about outcomes opposite to targets (see red boxes in the figure). However, our pattern a) introduces poses from further future to learn a PK representation by interpolating the in-between poses during the first step, and b) extrapolates (i.e. predicts) the sequence given historical poses only, but with PK to be distilled that offers supervision on the process of encoding and decoding during the second step.}
   \label{fig-intro}
\end{figure}

Classical sequential networks such as Recurrent Neural Networks (RNNs) \cite{5-2,18,26-1} and Long Short-Term Memory (LSTM) \cite{5-1,5-3,21} have been used to tackle certain sequence prediction frame-by-frame, yet suffer from error accumulation and convergence to the static mean pose \cite{7-1,7-2}. Temporal Convolutional Networks (TCNs) \cite{8-1,8-2} and the Discrete Cosine Transform (DCT) \cite{27,41,53} method have also proved their ability to encode temporal information, and to be coordinated with GCNs to better predict human motions in terms of both spatial and temporal domain \cite{27,29,41,53}. All the methods mentioned above follow the same prediction pattern, which is building a mapping relation between the sequence observed and the one to be predicted. However, it still remains tough to directly extrapolate a multivariate time series based on the historical sequence, and unexpected results or unreasonable poses may arise.

Inspired by the fact that, generally, \textbf{sequence interpolation is easier to operate and often yields an overall better result than extrapolation}, we introduce the overlooked poses which exist after the predicted sequence, in the hope of constructing a prediction pattern that \textbf{shares similar spirit with interpolation}. As these poses are only observable for interpolation and should not appear in the final prediction, we regard the information provided by them as privileged information. The concept of learning with privileged information is first presented by Vapnik \emph{et al}. \cite{57} in which the additional information of training samples can be used during training but discarded while testing. As is shown in the top half part of our prediction pattern in Figure \ref{fig-intro}, when the privileged sequence is introduced, the poses to be predicted finally are first interpolated as we hope. With this preliminary step done, another problem that arises is how to let the prediction step acquire the privileged information, since the raw privileged sequence would not exist then.

Recently, it has been proved that knowledge distillation is able to transfer knowledge between different models, and has been applied to diverse fields such as image dehazing \cite{14}, object detection \cite{16}, and online action detection \cite{11}. This advancement enlightens us to tackle the question above by designing a novel prediction network (see the bottom half of our pattern in Figure \ref{fig-intro}), which distills privileged knowledge (PK) from the aforementioned interpolation period. The distillation process exactly reflects the effect of PK supervision, enabling the model to be armed with a certain amount of PK gradually without taking the privileged sequence as input. In this way, this model is able to \textbf{keep the merits of interpolation}, while \textbf{predicting in the standard manner of extrapolation} finally.

\begin{figure*}[t]
  \centering
   \includegraphics[width=1.0\linewidth]{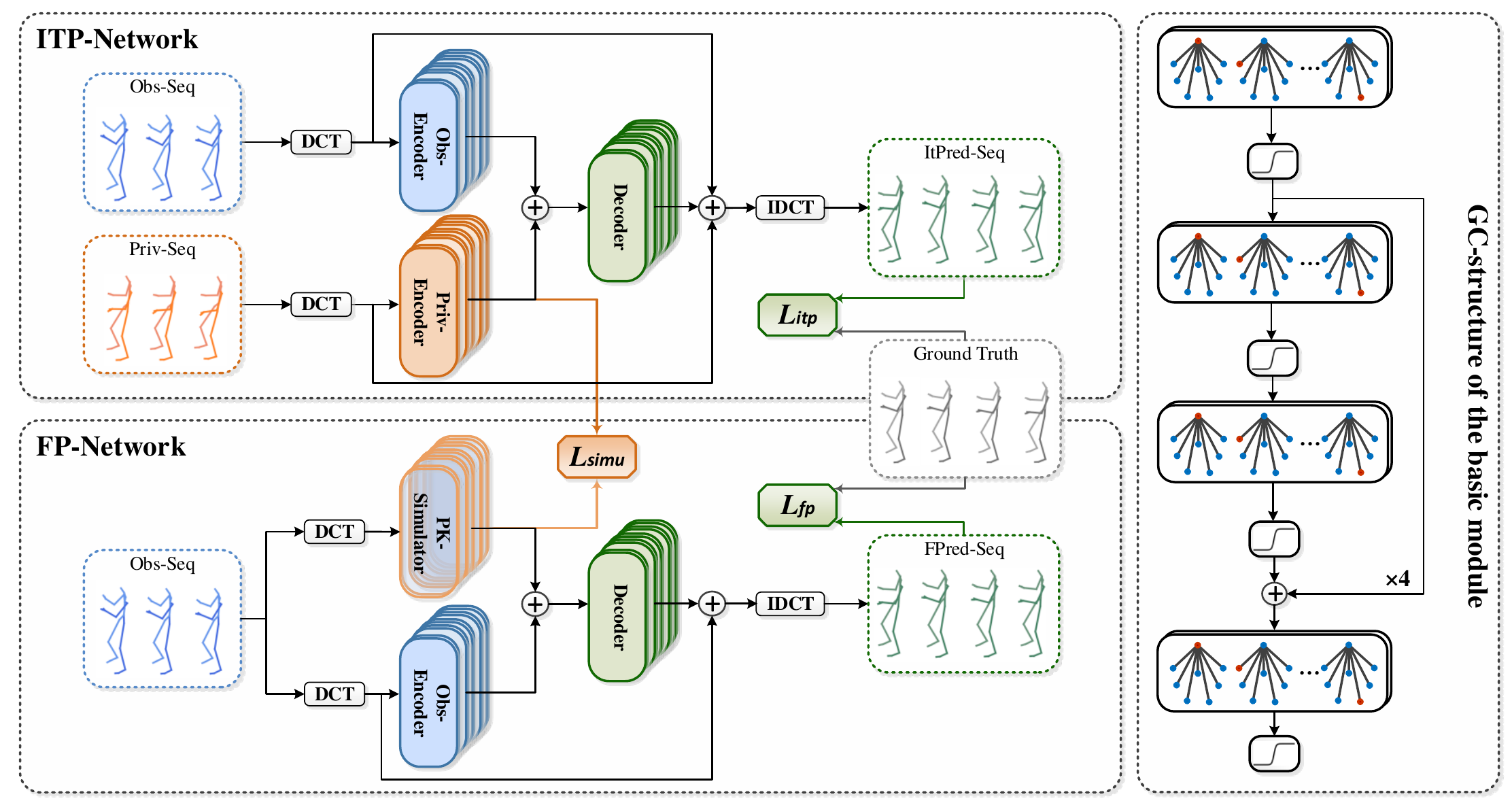}

   \caption{The backbone of our PK-GCN. ITP is composed of an Observed-sequence-Encoder (Obs-Encoder), a Privileged-sequence-Encoder (Priv-Encoder) and a decoder to interpolate the in-between sequence. FP is composed of an Obs-Encoder, a PK-Simulator and a decoder. These basic modules share the same architecture of graph convolutional layers (GC-Layers) and residual structures shown in the right sub-figure. During the interpolation period, a PK representation is learned by Priv-Encoder, so that the PK-Simulator is able to distill PK by approximating this representation (we measure their similarity by PK simulation loss $ L_{\rm simu} $). Therefore, FP could take as input only the observed sequence but predict motions with PK assisted.}
   \label{fig-pk-gcn}
\end{figure*}

In this paper, we name the two steps of networks proposed above as InTerPolation learning Network (ITP-Network) and Final Prediction Network (FP-Network), respectively. As our networks are created on the basis of GCN structure, the entire model is named as PK-GCN (see Figure \ref{fig-pk-gcn}). This model embodies a new pattern of human motion prediction, instead of the traditional one that directly finds an extrapolation relation between the observed sequence and the one to be predicted. Specifically, the Privileged-sequence-Encoder (Priv-Encoder) in ITP learns a PK representation of the privileged sequence during interpolation, then the novel PK-Simulator embedded in FP implements the PK distillation process by approximating the PK representation, with only the observed sequence as input. Meanwhile, a PK simulation loss function is presented to measure this approximation.

The idea of introducing future information to provide additional supervision has been practiced in prediction-related tasks, such as using predicted future frames to assist online action detection \cite{58}, or encapsulating future information in observation representations with Jaccard similarity measures \cite{56} for action anticipation, among which the information obtained from \emph{future} are still limited to the traditional observed-predicted research window. However, 3D human motion data is more elaborated with abundant yet subtle spatio-temporal information, and focusing on this two-section research window is not enough. Compared with existing works, our novelty lies in that we extend it to the observed-predicted-privileged window, in the seek of additional yet tailored information to provide supervision. To the best of our knowledge, this is the first work to involve poses existing after the predicted sequence to learn PK for human motion prediction.

In summary, our contributions are as follows: (i) ITP-Network: introducing the privileged sequence to learn PK during interpolation period; (ii) FP-Network: with the novel PK-Simulator distilling PK, to predict in the extrapolation manner while keeping the merits of interpolation; (iii) Entire model of PK-GCN: SOTA performance on benchmarked H3.6M, CMU-Mocap and 3DPW datasets.

\section{Related Work}
\label{sec: relatedwork}

\subsection{Human Motion Prediction}
\label{subs: HMPred}

Since human motion data is composed of a time series of human poses, many works on human motion prediction rely on sequential networks such as RNNs \cite{5-2,18,26-1}, LSTM \cite{5-1,21,guided} or GRU \cite{21,quaternion,tnn} to extract temporal information, and has achieved success. Notably, there occurs a problem that the first predicted frame is discontinuous with the last observed frame, so Martinez \emph{et al}. \cite{21} propose a residual single-layer GRU model which moves from the usual LSTM architectures \cite{22} and predicts velocities rather than poses. Meanwhile, a zero-velocity baseline is proposed in this work to prove that the repetition of the last observed pose even outperform prediction performances in\cite{5-1,5-2}. However, the crucial defect of RNNs, error accumulation, may lead to failure cases of convergence to unexpected mean poses. Other classical architectures such as CNN-based \cite{7-1,trajectorycnn} make it possible to predict in a sequence-to-sequence manner rather than frame-by-frame, while GAN-based \cite{7-2,54} are aimed at generating more realistic motions.

Recently, GCNs gain huge popularity to model spatial dependencies on predict human motions \cite{27,41,29,26-2,30-2,60,cues,progressively,gating}. Mao \emph{et al}. \cite{27} first introduce a generic graph instead of the previous skeletal kinematic tree \cite{28}, with DCT encoding the short-term history in frequency domain. To depict the topological relations of body joints, Cui \emph{et al}. \cite{29} present a dynamic GCN, which learns both the natural connectivity of joint pairs and the implicit links of geometrically separated joints. Different multi-scale GNN \cite{30-1}/GCN \cite{30-2} models are developed to further explore the features of human motions. Sofianos \emph{et al}. \cite{60} realizes cross-talk over space and time within one graph and factorizes it into separable matrices for better understanding on both features. Meanwhile, stochastic motion prediction \cite{52,59,53,mao2022,KD-hmp} gradually draw attention, which generate diverse yet plausible motion sequences given only one observed.

All the models mentioned above follow the traditional prediction pattern, limiting their research window within the observed-predicted sequence. Differently, our work goes further by adding a privileged window containing overlooked poses, so that a PK representation can be learned and distilled through our two-step model, to serve as a supplement of supervision information to assist prediction tasks.

\subsection{Knowledge Distillation}
\label{subs: KD}

Early knowledge distillation is utilized for deep model compression \cite{12}, wherein the knowledge from the larger teacher network helps the smaller student network to learn better \cite{13}. The teacher provides the soft target for the student to imitate the output logits and intermediate features \cite{32-1,32-2,31-3,31-4}. Subsequently, it has been proved that knowledge distillation is able to transfer knowledge between different models, and has its applications in image classification \cite{33,32-1}, action detection \cite{11}, semantic segmentation \cite{34}, trajectory prediction \cite{KD-tp} and stochastic motion prediction \cite{KD-hmp}. However, the teachers in these models provides additional information about training samples either with no privileged information involved, or with it involved but irrelevant to distill knowledge from future further away. In addition, our design of distillation process ensures that a) PK would be distilled into FP as much as possible, and b) only the observed sequence would be taken as input. We embed the PK-Simulator in FP to be fed with the observed sequence but approximate the PK representation learned by Priv-Encoder, and present a PK simulation loss $ L_{\rm simu} $ (also denoted as PKSL) to measure this approximation.

\section{Proposed Method}
\label{sec: propmeth}

We present a new formulation for human motion prediction. Different from existing works, our notation involves three research windows (i.e. the observed-predicted-privileged sequence), with $ N $ poses $ \textbf{X}_{1:N}=[\textbf{x}_1, \textbf{x}_2, \dots, \textbf{x}_N] $ given, $ T $ poses $ \textbf{X}_{N+1:N+T} $ to predict, and particularly another $ P $ poses after the predicted sequence $ \textbf{X}_{N+T+1:N+T+P} $ as the privileged sequence, where $ \textbf{x}_i \in \mathbb{R}^K $, with $ K $ parameters depicting each pose based on joint coordinates or angles. The entire architecture of PK-GCN is shown in Figure \ref{fig-pk-gcn}. To be specific, ITP-Network encodes both the observed sequence and the privileged sequence to interpolate the predicted sequence, during which PK is learned and to be distilled to the next step, enabling FP-Network to predict in the standard manner of extrapolation finally, but with PK assisted.

\subsection{Spatio-Temporal Encoding}
\label{subs: st-encoding}

To encode both the spatial and temporal information of human motions, we employ the remarkable encoding method from \cite{27}. We treat each human skeleton as a fully-connected graph of $ K $ nodes, and use an adjacency matrix $ A \in \mathbb{R} ^{K \times K} $ to depict its edges. Formally, we define $ H^{(l)} \in \mathbb{R} ^{K \times F^{(l)}} $ to be fed into the $ l $th layer of GCN (i.e. the $ l $th GC-Layer), $ W^{(l)} \in \mathbb{R} ^{ F^{(l)} \times F^{(l+1)} } $ as weight matrix, and output (also the input of next GC-Layer) as:
\begin{equation}
  H^{(l+1)}=\sigma(A^{(l)}H^{(l)}W^{(l)}),
  \label{gc-layer}
\end{equation}
where $ \sigma(\cdot) $ is an activation function. Particularly, a DCT-based representation is employed to extract temporal information, which transforms the observed sequence into its DCT coefficients $ H^{(1)} \in \mathbb{R} ^{K \times C} $ before being input into our network, with $ C $ the DCT coefficient number. The output is $ H^{(last)} \in \mathbb{R} ^{K \times C} $, and should also be transformed into its sequence form by IDCT.

\subsection{Network Structure}

\textbf{The InTerPolation learning Network (ITP-Network)}, based on residual GCN architecture, is composed of two encoders to encode the observed sequence (Obs-Encoder) and the privileged sequence (Priv-Encoder), and one decoder to interpolate the in-between sequence (see the top half of Figure \ref{fig-pk-gcn}). In each of the encoders/decoder, we employ 4 residual blocks and 2 separate GC-Layers with one at the beginning and one at the end (see the right part of Figure \ref{fig-pk-gcn}). Each residual block is composed of 2 GC-Layers, and the activation function is set to $ tanh(\cdot) $. The outputs of the Obs-Encoder and the Priv-Encoder are integrated by element-wise addition.

It should be emphasized that PK \emph{only} serves as a supplement of supervision information, and it is unreasonable to hold this advantage as an excuse to diminish the original predictability of Obs-Encoder in FP. Additionally, overly complicated PK is not conducive to distill. Therefore, in this PK preparation step, we intend to reduce the relative weight of Priv-Encoder. However, this variation will lead to degradation on the interpolation performance, so we alternatively set the weight ratio of the outermost residual connection to 0.7:0.3, sum of which is 1 in line with the residual weight in FP.

Since the input of both encoders and the output of the decoder are all DCT coefficients of the corresponding sequences, motivated by zero-velocity baseline \cite{21} and inspired by the padding operation in \cite{27}, we propose a new data pre-processing method which replicates the last pose of the observed sequence, $ \textbf{x}_N $, $ T+P $ times (represented as $ \textbf{ObsX}_{1:N+T+P} $), and the first pose of the privileged sequence, $ \textbf{x}_{N+T+1} $, $ N+T $ times (represented as $ \textbf{PrivX}_{1:N+T+P} $), to form residual vectors in frequency domain as the input $ H_{\rm obs}^{(1)} $, $ H_{\rm priv}^{(1)} $. After epochs of training to find the interpolation relation $ \mathcal{I} $:
\begin{equation}
  H_{\rm itp}^{(last)} = \mathcal{I}(H_{\rm obs}^{(1)}, H_{\rm priv}^{(1)}),
  \label{itp-network}
\end{equation}
where the interpolation result $ H_{\rm itp}^{(last)} $ is close to the ground truth, the Priv-Encoder could obtain its most accurate representation of PK, which is aimed to be distilled to the final prediction period.

\textbf{The Final Prediction Network (FP-Network)} consists of an Obs-Encoder to encode the observed sequence, a novel PK-Simulator to approximate the representation obtained from Priv-Encoder, and a decoder to extrapolate the predicted sequence. We construct each module the same as in ITP with residual blocks and GC-Layers. We pad the observed sequence to the length of $ N+T+P $ the same as in ITP, and operate DCT transformation to form the input $ H_{\rm obs}^{(1)} $ to be fed into the network, in the seek of a predictive function $ \mathcal{P} $:
\begin{equation}
  H_{\rm fp}^{(last)} = \mathcal{P}(H_{\rm obs}^{(1)}),
  \label{fp-network}
\end{equation}
with distilled PK assisted. Since the privileged sequence does not exist at this moment, the embedded PK-Simulator should undertake the assistant role by distilling PK but without direct use of raw privileged data. A PK simulation loss is introduced in this step, to measure the discrepancy between the output of PK-Simulator and Priv-Encoder (i.e. how well does PK-Simulator imitate the PK representation learned by Priv-Encoder), and notably, the PK representation learned by Priv-Encoder has been well-trained previously and should remain fixed. Moreover, the distillation process is not damage-free. The PK-Simulator is doomed not to obtain entire PK from ITP, but is still capable of reflecting PK to some extent.

\subsection{Training}
\label{subs: Training}

Human poses are depicted mainly based on their 3D coordinates (i.e. 3D positions) or angles of body joints. In this section, we follow \cite{27,41,30-2,60} to employ Mean Per Joint Position Error (MPJPE) for position-based representation, and Mean Angle Error (MAE) \cite{21,27,60} for angle-based representation.

\subsubsection{Loss for ITP-Network.}
\label{subsubs: loss-itp}

The MPJPE interpolation loss is expressed as:
\begin{equation}
  L_{{\rm itp(mp)}}=\frac{1}{(N+T+P)J}\sum_{n = 1}^{N+T+P}\sum_{j = 1}^{J}\left\lVert \hat{\textbf{m}}_{j,n}-\textbf{m}_{j,n} \right\rVert _{2},
  \label{eq:L_itp-mpjpe}
\end{equation}
where $ \hat{\textbf{m}}_{j,n} \in \mathbb{R}^3 $ is the interpolated $ j $th joint position of the pose $ \textbf{x}_n $, and $ \textbf{m}_{j,n} $ is the corresponding ground truth one. Note that the length of the interpolated sequence should be consistent with the padded observed/privileged sequences (i.e. $ N+T+P $), and is denoted as $ \textbf{ItPredX}_{1:N+T+P} $. This design is inspired by \cite{27}, which calculates the sum of $ l_2 $ over the entire sequence sample, and we extend it to the longer observed-predicted-privileged window to fit our privileged design. Meanwhile, the minimization on $ L_{{\rm itp(mp)}} $ ensures that the optimal PK representation is obtained by Priv-Encoder, denoted as $ \textbf{E}_m $.

Similarly, the MAE interpolation loss is expressed as:
\begin{equation}
  L_{\rm itp(ma)}=\frac{1}{(N+T+P)K}\sum_{n = 1}^{N+T+P}\sum_{k = 1}^{K}\left\lVert \hat{\textbf{a}}_{k,n}-\textbf{a}_{k,n} \right\rVert _{1},
  \label{eq:L_itp-mae}
\end{equation}
where $ \hat{\textbf{a}}_{k,n} $ is the interpolated $ k $th angle of the pose $ \textbf{x}_n $, and $ \textbf{a}_{k,n} $ is the corresponding ground truth one. We minimize $ L_{\rm itp(ma)} $ to gain the optimal output of the Priv-Encoder represented as $ \textbf{E}_a $.

\subsubsection{Loss for FP-Network.}
\label{subsubs: loss-fp}

As FP is aimed at imitating the interpolation process, then the input form should maintain the length of $ N+T+P $ in previous step, and the predicted sequence is also the same, denoted as $ \textbf{FPredX}_{1:N+T+P} $. But notably, we only take into account the loss of $ N+T $ poses because the following $ P $ ones are not our prediction target, and the corresponding ground truth (i.e. the privileged sequence) is not observable for this step. To measure the discrepancy between the predicted sequence and its ground truth, we present MPJPE prediction loss
\begin{equation}
  L_{\rm fp(mp)}=\frac{1}{(N+T)J}\sum_{n = 1}^{N+T}\sum_{j = 1}^{J}\left\lVert \hat{\textbf{m}}_{j,n}-\textbf{m}_{j,n} \right\rVert _{2}
  \label{eq:L_fp-mpjpe}
\end{equation}
for position-based joint representation and MAE prediction loss
\begin{equation}
  L_{\rm fp(ma)}=\frac{1}{(N+T)K}\sum_{n = 1}^{N+T}\sum_{k = 1}^{K}\left\lVert \hat{\textbf{a}}_{k,n}-\textbf{a}_{k,n} \right\rVert _{1}
  \label{eq:L_fp-mae}
\end{equation}
for angle-based one. Meanwhile, we denote the output of the PK-Simulator as $ \textbf{S}_m $ for MPJPE and $ \textbf{S}_a $ for MAE, which approximate $ \textbf{E}_m $ and $ \textbf{E}_a $ respectively to implement distillation process. The PK simulation loss $ L_{\rm simu} $ (i.e. PKSL) we introduce to measure this approximation is formulated as:
\begin{equation}
  L_{\rm simu(mp)}=\left\lVert \textbf{S}_m-\textbf{E}_m\right\rVert_F \quad \quad {\rm or} \quad \quad L_{\rm simu(ma)}=\left\lVert \textbf{S}_a-\textbf{E}_a\right\rVert_F.
  \label{eq:L_simu}
\end{equation}
Therefore, the MPJPE/MAE total loss are:
\begin{equation}
  L_{\rm total(mp)}=L_{\rm fp(mp)}+\lambda L_{\rm simu(mp)} \quad {\rm or} \quad L_{\rm total(ma)}=L_{\rm fp(ma)}+\lambda L_{\rm simu(ma)},
  \label{eq:L_total}
\end{equation}
where $ \lambda $ is the relative weight of the final prediction error and the simulation error, and is set to 0.6.

\section{Experiments}
\label{sec: Experiments}

To evaluate our proposed model, we conduct experiments on benchmarked motion capture datasets, including Human3.6M (H3.6M), CMU-Mocap, and 3DPW datasets. In this section, we first introduce these datasets, the evaluation metrics, and the baselines which we compare with, and then present our results.

\subsection{Datasets and Evaluation Metrics}
\label{subs: datasets&metrics}

\noindent\textbf{Human3.6M} (H3.6M) \cite{37}, depicting seven actors which perform 15 actions (walking, eating, smoking, discussing, etc.), is the most widely used dataset for human motion processing tasks. Each human pose is represented by 32 joints to form a skeleton. In line with previous methods \cite{21,7-1,7-2,27}, we set the global rotation and translation to zero and down-sample the sequences to 25 frames per second. We train our model on Subject 1 (S1), S6, S7, S8, S9 and test on S5 the same as \cite{21,7-1,7-2,27}.

\noindent\textbf{CMU-Mocap}. Based on \cite{7-1}, eight actions are selected for experiments after removing sequences with multiple actors, insufficient poses, or actions with repetitions. The other pre-processing operations are the same as on H3.6M.

\noindent\textbf{3DPW}. The 3D Pose in the Wild dataset \cite{3dpw} consists of 60 sequences about both indoor and outdoor actions captured by a moving camera at 30 fps. We follow the setting in \cite{27} to evaluate the effectiveness of our models on challenging activities.

\noindent\textbf{Evaluation Metrics.} We employ Mean Per Joint Position Error (MPJPE) \cite{37} in millimeter \cite{27,30-2,60} to measure the position discrepancy between the interpolated/predicted joint and the corresponding ground truth, and Mean Angle Error (MAE) \cite{21,27,60} to measure the angle one. As is stated in \cite{27} that MAE may fail due to ambiguous representation, we focus mainly on MPJPE results.

\begin{table*}[ht]
  \caption{Comparisons of short-term MPJPE error on H3.6M dataset. The best result is highlighted in bold.}
  \label{h36-3d-short}
  \centering
  \renewcommand{\arraystretch}{1.2}
  \setlength\tabcolsep{3.2pt}
  \scalebox{0.57}{
    \begin{tabular}{c|cccc|cccc|cccc|cccc|cccc}
      & \multicolumn{4}{c|}{walking} & \multicolumn{4}{c|}{eating} & \multicolumn{4}{c|}{smoking} & \multicolumn{4}{c|}{discussion} & \multicolumn{4}{c}{directions} \\
     millisecond (ms) & 80 & 160 & 320 & 400 & 80 & 160 & 320 & 400 & 80 & 160 & 320 & 400 & 80 & 160 & 320 & 400 & 80 & 160 & 320 & 400 \\ \hline
     Res. sup \cite{21} & 29.4 & 50.8 & 76.0 & 81.5 & 16.8 & 30.6 & 56.9 & 68.7 & 23.0 & 42.6 & 70.1 & 82.7 & 32.9 & 61.2 & 90.9 & 96.2 & 35.4 & 57.3 & 76.3 & 87.7 \\
     Traj-GCN \cite{27} & 12.3 & 23.0 & 39.8 & 46.1 & 8.4 & \textbf{16.9} & 33.2 & 40.7 & 8.0 & 16.2 & 31.9 & 38.9 & 12.5 & 27.4 & 58.5 & 71.7 & 9.0 & 19.9 & 43.4 & \textbf{53.7} \\
     MSR-GCN \cite{30-2} & 12.2 & 22.7 & 38.6 & 45.2 & 8.4 & 17.1 & \textbf{33.0} & \textbf{40.4} & 8.0 & 16.3 & 31.3 & 38.2 & 12.0 & 26.8 & 57.1 & 69.7 & 8.6 & \textbf{19.7} & \textbf{43.3} & 53.8 \\
     STS-GCN \cite{60} & 16.7 & 27.1 & 43.4 & 51.0 & 11.7 & 18.9 & 35.2 & 43.3 & 12.3 & 19.8 & 35.6 & 42.3 & 15.3 & 28.1 & 58.5 & 71.9 & 12.8 & 23.7 & 51.1 & 63.2 \\ \hline
     PK-GCN & \textbf{8.9} & \textbf{15.9} & \textbf{28.0} & \textbf{31.6} & \textbf{8.1} & 17.7 & 33.6 & 41.8 & \textbf{7.4} & \textbf{14.3} & \textbf{24.4} & \textbf{29.2} & \textbf{10.3} & \textbf{22.9} & \textbf{42.0} & \textbf{47.2} & \textbf{8.6} & 23.7 & 46.5 & 56.2 \\ \hline
      & \multicolumn{4}{c|}{greeting} & \multicolumn{4}{c|}{phoning} & \multicolumn{4}{c|}{posing} & \multicolumn{4}{c|}{purchases} & \multicolumn{4}{c}{sitting} \\
     millisecond (ms) & 80 & 160 & 320 & 400 & 80 & 160 & 320 & 400 & 80 & 160 & 320 & 400 & 80 & 160 & 320 & 400 & 80 & 160 & 320 & 400 \\ \hline
     Res. sup \cite{21} & 34.5 & 63.4 & 124.6 & 142.5 & 38.0 & 69.3 & 115.0 & 126.7 & 36.1 & 69.1 & 130.5 & 157.1 & 36.3 & 60.3 & 86.5 & 95.9 & 42.6 & 81.4 & 134.7 & 151.8 \\
     Traj-GCN \cite{27} & 18.7 & 38.7 & 77.7 & 93.4 & 10.2 & 21.0 & 42.5 & 52.3 & 13.7 & 29.9 & 66.6 & 84.1 & 15.6 & 32.8 & 65.7 & 79.3 & 10.6 & \textbf{21.9} & \textbf{46.3} & 57.9 \\
     MSR-GCN \cite{30-2} & 16.5 & 37.0 & 77.3 & 93.4 & \textbf{10.1} & 20.7 & 41.5 & 51.3 & 12.8 & 29.4 & 67.0 & 85.0 & 14.8 & 32.4 & 66.1 & 79.6 & 10.5 & 22.0 & \textbf{46.3} & 57.8 \\
     STS-GCN \cite{60} & 18.7 & 34.9 & 71.6 & 86.4 & 13.7 & 22.4 & 43.6 & 53.8 & 16.4 & 30.4 & 67.6 & 84.7 & 19.1 & 35.8 & 70.2 & 83.1 & 15.2 & 25.1 & 49.8 & 60.8 \\ \hline
     PK-GCN & \textbf{13.3} & \textbf{27.2} & \textbf{67.3} & \textbf{83.1} & 11.4 & \textbf{20.2} & \textbf{37.7} & \textbf{43.2} & \textbf{9.1} & \textbf{23.6} & \textbf{65.8} & \textbf{81.2} & \textbf{15.2} & \textbf{31.4} & \textbf{57.9} & \textbf{68.0} & \textbf{10.1} & 24.6 & 47.8 & \textbf{57.3} \\ \hline
      & \multicolumn{4}{c|}{sittingdown} & \multicolumn{4}{c|}{takingphoto} & \multicolumn{4}{c|}{waiting} & \multicolumn{4}{c|}{walkingdog} & \multicolumn{4}{c}{walkingtogether} \\
     millisecond (ms) & 80 & 160 & 320 & 400 & 80 & 160 & 320 & 400 & 80 & 160 & 320 & 400 & 80 & 160 & 320 & 400 & 80 & 160 & 320 & 400 \\ \hline
     Res. sup \cite{21} & 47.3 & 86.0 & 145.8 & 168.9 & 26.1 & 47.6 & 81.4 & 94.7 & 30.6 & 57.8 & 106.2 & 121.5 & 64.2 & 102.1 & 141.1 & 164.4 & 26.8 & 50.1 & 80.2 & 92.2 \\
     Traj-GCN \cite{27} & 16.1 & 31.1 & 61.5 & 75.5 & 9.9 & 20.9 & 45.0 & 56.6 & 11.4 & 24.0 & 50.1 & 61.5 & 23.4 & 46.2 & 83.5 & 96.0 & 10.5 & 21.0 & 38.5 & 45.2 \\
     MSR-GCN \cite{30-2} & 16.1 & 31.6 & 62.5 & 76.8 & 9.9 & 21.0 & 44.6 & 56.3 & 10.7 & 23.1 & \textbf{48.3} & \textbf{59.2} & \textbf{20.7} & 42.9 & \textbf{80.4} & \textbf{93.3} & 10.6 & 20.9 & 37.4 & \textbf{43.9} \\
     STS-GCN \cite{60} & 22.1 & 37.2 & 66.5 & 79.4 & 14.5 & 23.4 & 47.8 & 59.4 & 14.5 & 24.8 & 50.4 & 62.0 & 26.5 & 46.1 & 83.6 & 97.3 & 14.3 & 23.9 & 41.3 & 49.1 \\ \hline
     PK-GCN & \textbf{11.5} & \textbf{27.5} & \textbf{56.8} & \textbf{67.3} & \textbf{7.6} & \textbf{16.1} & \textbf{39.7} & \textbf{51.3} & \textbf{9.5} & \textbf{23.0} & 55.9 & 63.6 & 21.3 & \textbf{42.4} & 83.7 & 95.1 & \textbf{9.4} & \textbf{19.3} & \textbf{36.3} & 44.8 \\ \hline
     \end{tabular}}
\end{table*}

\begin{table*}[ht]
  \caption{Comparisons of long-term MPJPE error on some actions in H3.6M dataset. The best result is highlighted in bold.}
  \label{h36-3d-long}
  \centering
  \renewcommand{\arraystretch}{1.1}
  \setlength\tabcolsep{3.2pt}
  \scalebox{0.76}{
    \begin{tabular}{c|cc|cc|cc|cc|cc|cc}
      & \multicolumn{2}{c|}{walking} & \multicolumn{2}{c|}{eating} & \multicolumn{2}{c|}{smoking} & \multicolumn{2}{c|}{discussion} & \multicolumn{2}{c|}{directions} & \multicolumn{2}{c}{greeting} \\
     millisecond (ms) & 560 & 1000 & 560 & 1000 & 560 & 1000 & 560 & 1000 & 560 & 1000 & 560 & 1000 \\ \hline
     Res. sup \cite{21} & 81.7 & 100.7 & 79.9 & 100.2 & 94.8 & 137.4 & 121.3 & 161.7 & 110.1 & 152.5 & 156.1 & 166.5 \\
     Traj-GCN \cite{27} & 54.1 & 59.8 & 53.4 & 77.8 & 50.7 & 72.6 & 91.6 & 121.5 & \textbf{71.0} & 101.8 & 115.4 & 148.8 \\
     MSR-GCN \cite{30-2} & 52.7 & 63.0 & \textbf{52.5} & 77.1 & 49.5 & 71.6 & 88.6 & 117.6 & 71.2 & \textbf{100.6} & 116.3 & 147.2 \\
     STS-GCN \cite{60} & 58.0 & 70.2 & 57.4 & 82.6 & 55.5 & 76.1 & 91.1 & 118.9 & 79.9 & 109.6 & 106.3 & 136.1 \\ \hline
     PK-GCN & \textbf{42.5} & \textbf{47.0} & 57.9 & \textbf{69.7} & \textbf{33.3} & \textbf{60.2} & \textbf{75.8} & \textbf{112.0} & 74.7 & 101.9 & \textbf{90.1} & \textbf{108.8} \\ \hline
     \end{tabular}}
\end{table*}

\subsection{Baselines}
\label{subs: baselines}

Recent methods are involved to evaluate the effectiveness of our model, denoted as Res.sup. \cite{21}, Traj-GCN \cite{27}, MSR-GCN \cite{30-2} and STS-GCN \cite{60}. Specifically, Traj-GCN \cite{27} is the first work that introduces an unconstrained graph convolutional network on human motion prediction. MSR-GCN \cite{30-2} presents a GCN with multi-scale residual structure which abstracts human poses on different levels. STS-GCN \cite{60} realizes cross-talk of space and time within only one graph, and is able to factorize the graph into separable space and time matrices for better understanding of both features. We follow them to predict 10 (i.e. short-term) or 25 (i.e. long-term) poses with the observed 10 poses. It should be added that errors in \cite{21,27,30-2} are shown by per frame, while the one in \cite{60} is shown by averaging on all previous frames. To keep in line with Traj-GCN and MSR-GCN, we re-demonstrate the error in \cite{60} as the same protocol in \cite{21,27,30-2}.

\subsection{Implement Details}
\label{subs: imdetails}

We implement our model using PyTorch \cite{48} on an NVIDIA 2080TI GPU. The optimizer is ADAM \cite{49}, and the learning rate is set to 0.0005 with a 0.96 decay every two epochs. Each GC-Layer has a dropout rate of 0.5, with the learnable weight matrix $ \textbf{W} $ of size $ 256\times256 $. The batch size is set to 16. The ITP-Network and the FP-Network are trained separately. We first train the former one for 50 epochs, then fix it and train the latter for another 50 epochs. The size of our model is 2.8M for both stages.

\begin{table*}[ht]
  \caption{Comparisons of average MPJPE error (left) and MAE error (right) on H3.6M dataset. The best result is highlighted in bold.}
  \label{h36-average}
  \centering
  \renewcommand{\arraystretch}{1.1}
  \setlength\tabcolsep{3.2pt}
  \scalebox{0.76}{
    \begin{tabular}{c|cccccc||cccccc}
      & \multicolumn{6}{c||}{average-MPJPE} & \multicolumn{6}{c}{average-MAE} \\
     millisecond (ms) & 80 & 160 & 320 & 400 & 560 & 1000 & 80 & 160 & 320 & 400 & 560 & 1000 \\ \hline
     Res. sup \cite{21} & 34.7 & 62.0 & 101.1 & 115.5 & 135.8 & 167.3 & 0.36 & 0.67 & 1.02 & 1.15 & - & - \\
     Traj-GCN \cite{27} & 12.7 & 26.1 & 52.3 & 63.5 & 81.6 & 114.3 & 0.32 & 0.55 & 0.91 & 1.04 & 1.27 & 1.66 \\
     MSR-GCN \cite{30-2} & 12.1 & 25.6 & 51.6 & 62.9 & 81.1 & 114.2 & - & - & - & - & - & - \\
     STS-GCN \cite{60} & 16.3 & 28.1 & 54.4 & 65.8 & 85.1 & 117.0 & 0.31 & 0.57 & 0.91 & 1.03 & 1.22 & 1.61 \\ \hline
     PK-GCN & \textbf{10.8} & \textbf{23.3} & \textbf{48.2} & \textbf{57.4} & \textbf{76.1} & \textbf{106.4} & \textbf{0.29} & \textbf{0.54} & \textbf{0.85} & \textbf{0.96} & \textbf{1.15} & \textbf{1.57} \\ \hline
     \end{tabular}}
\end{table*}

\begin{figure}[ht]
  \centering
  \subfigure[]{
  \includegraphics[width=0.42\linewidth]{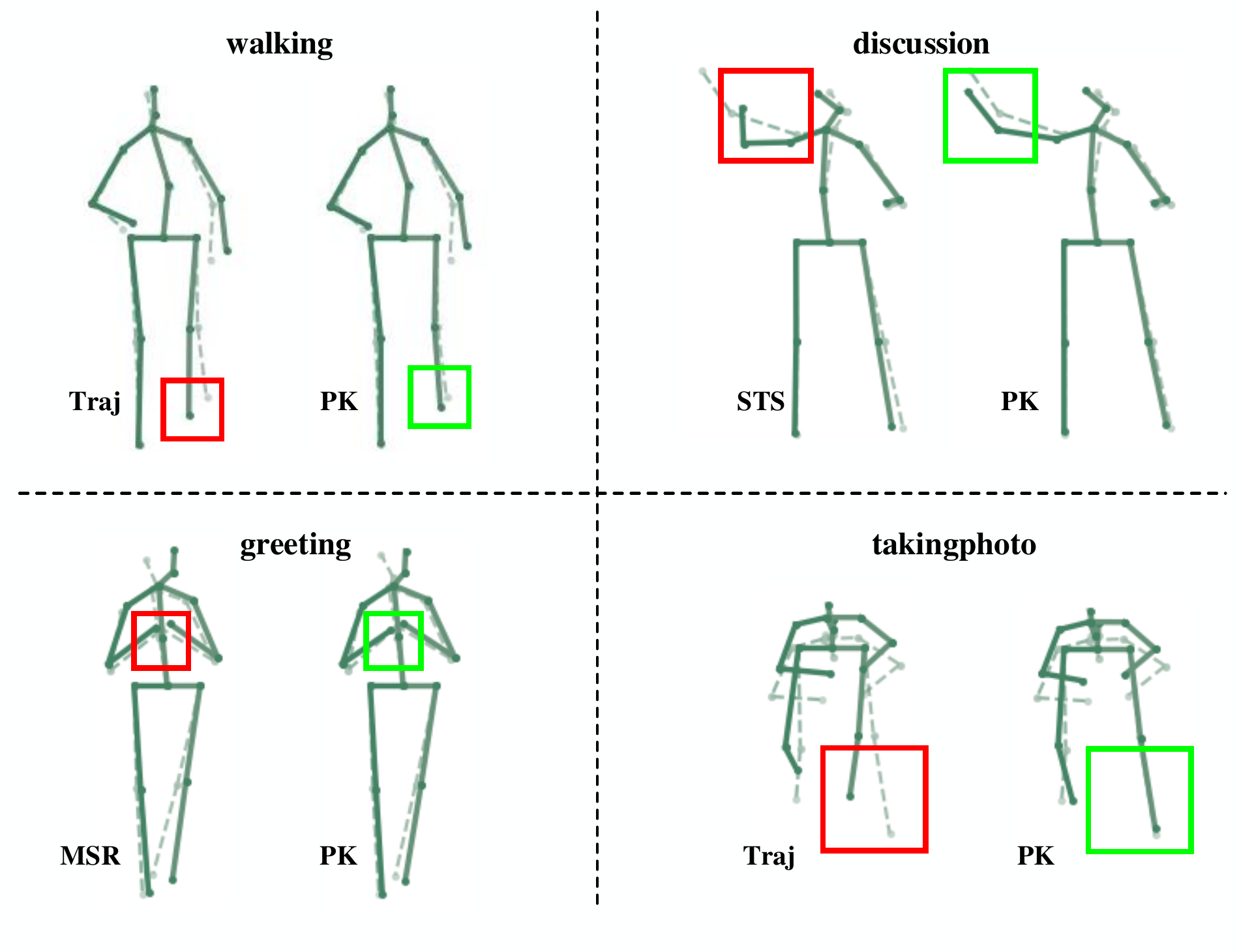}}
  \quad \quad
  \subfigure[]{
  \includegraphics[width=0.48\linewidth]{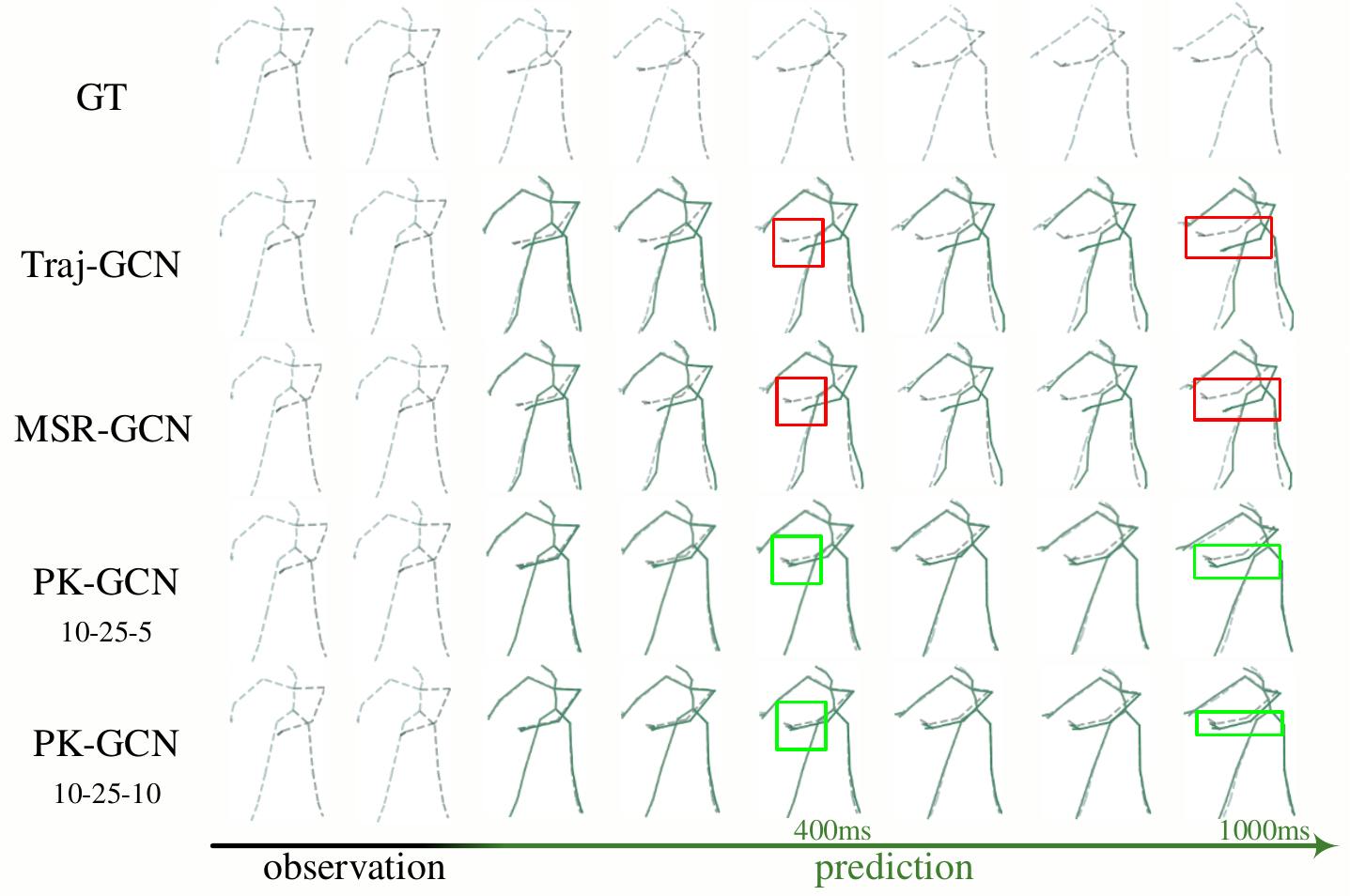}}
  \caption{(a) Visualized comparisons between three baselines \cite{27,30-2,60} and PK-GCN on actions \emph{walking}, \emph{discussion}, \emph{greeting}, \emph{takingphoto} in H3.6M at the 1000ms testpoint (i.e. 25th frame). Performances marked in green boxes are better than in red. The dashed lines denote ground truth. (b) Visualized comparisons of short-term and long-term predictions on a motion sequence \emph{Washwindow} in CMU-Mocap. We show the ground truth, Traj-GCN \cite{27}, MSR-GCN\cite{30-2}, ours with 5 privileged poses of PK, and ours with 10 privileged poses of PK. The prediction deviations on left wrist are marked with red/green boxes. Our results are much closer to the ground truth.}
  \label{viz}
\end{figure}

\subsection{Results}
\label{subs: results}

Following the baselines \cite{27,30-2,60}, we evaluate our prediction performance by showing the results of 400ms short-term and 1000ms long-term predictions on H3.6M, CMU-Mocap, and 3DPW with 10-pose sequence observed. The length of the privileged sequence offered to PK-GCN is also set to 10.

\textbf{H3.6M.} Table \ref{h36-3d-short} and Table \ref{h36-3d-long} show quantitative comparisons for MPJPE error between short/long-term prediction and the ground truth. We find that error obtained by PK-GCN is smaller than other baselines in most cases. Particularly, activities with larger ranges of motions, such as \emph{discussion} and \emph{takingphoto}, achieve more improvements than the others, in that the distilled PK restricts the overall moving trend/orientation of an activity from the future further away. Taking a sequence from \emph{greeting} as an example, when the person makes a bow with hands folded in front, the prediction trajectory of hands is forced to move from the chest to down with PK foretelling that the hands will be at down in the end. On the other hand, the failure of \emph{eating} and \emph{sitting} confirms our assumption that activities with relatively motionless status are less likely to benefit much from our model, as the information in observations is sufficient to generate a nearly static sequence.

\begin{table*}[ht]
  \caption{Comparisons of average MPJPE error of short/long-term prediction on CMU-Mocap (left) and 3DPW (right) datasets. The best results are highlighted in bold.}
  \label{cmu-3dpw}
  \centering
  \renewcommand{\arraystretch}{1.1}
  \setlength\tabcolsep{3.2pt}
  \scalebox{0.76}{
    \begin{tabular}{c|cccccc||ccccc}
      \multicolumn{1}{l|}{} & \multicolumn{6}{c||}{CMU-Average} & \multicolumn{5}{c}{3DPW-Average} \\
      millisecond (ms) & 80 & 160 & 320 & 400 & 560 & 1000 & 200 & 400 & 600 & 800 & 1000 \\ \hline
      Res. sup \cite{21} & 24.0 & 43.0 & 74.5 & 87.2 & 105.5 & 136.3 & 113.9 & 173.1 & 191.9 & 201.1 & 210.7 \\
      Traj-GCN \cite{27} & 11.5 & 20.4 & 37.8 & 46.8 & 55.8 & 86.2 & 35.6 & 67.8 & 90.6 & 106.9 & 117.8 \\
      MSR-GCN \cite{30-2} & \textbf{8.1} & 18.7 & 34.2 & 42.9 & 53.7 & 83.0 & - & - & - & - & - \\ \hline
      PK-GCN & 9.4 & \textbf{17.1} & \textbf{32.8} & \textbf{40.3} & \textbf{52.2} & \textbf{79.3} & \textbf{34.8} & \textbf{66.2} & \textbf{88.1} & \textbf{104.3} & \textbf{114.2} \\ \hline
      \end{tabular}}
\end{table*}

We further present the average MPJPE and MAE errors in Table \ref{h36-average}, wherein the omitted cells indicate no data source available. Meanwhile, visualization of prediction at 1000ms testpoint on four actions in H3.6M is given in Figure \ref{viz} (a).

\textbf{CMU-Mocap and 3DPW.} Similar experiments are conducted on CMU-Mocap and 3DPW datasets, and the MPJPE errors of which are shown in Table \ref{cmu-3dpw}. Apparently, our PK-GCN achieves state-of-the-art performance on CMU-Mocap among the baselines. In Figure \ref{viz} (b), qualitative visualization results of \emph{Washwindow} are demonstrated with the ground truth. The prediction results of Traj-GCN \cite{27}, MSR-GCN\cite{30-2}, our PK-GCN with 5 privileged poses of PK, and ours with 10 privileged poses of PK, are from the top to the bottom. These comparisons show that our model performs better than that of the baselines. It should be emphasized that the first and second numbers behind our model indicate the length of the observed sequence and the predicted sequence, while the third number $ n $ means that we train our ITP-Network by introducing $ n $ privileged poses to learn PK, and distill it to assist the final prediction. Errors in Table \ref{h36-3d-short}, \ref{h36-3d-long}, \ref{h36-average}, \ref{cmu-3dpw} are all obtained with PK of 10 poses.

In addition, the improvement achieved on 3DPW proves that PK is effective when faced with more complex data, but notably, it should be guaranteed that poses in one sample should be consistent. Two irrelevant actions in one sequence sample (such as \emph{arguing} to \emph{dancing}) is not suitable for PK distillation.

\subsection{Ablation Study}
\label{subs: ab-study}

In this section, we conduct more experiments on three parts for deeper analysis on our PK-GCN. Following the previous notation, any numbers of the form \emph{a-b-c} indicate predicting \emph{b} poses on \emph{a} historical poses, with \emph{c} poses of PK introduced.

\begin{figure}[ht]
  \centering
  \subfigure[]{
  \includegraphics[width=0.45\linewidth]{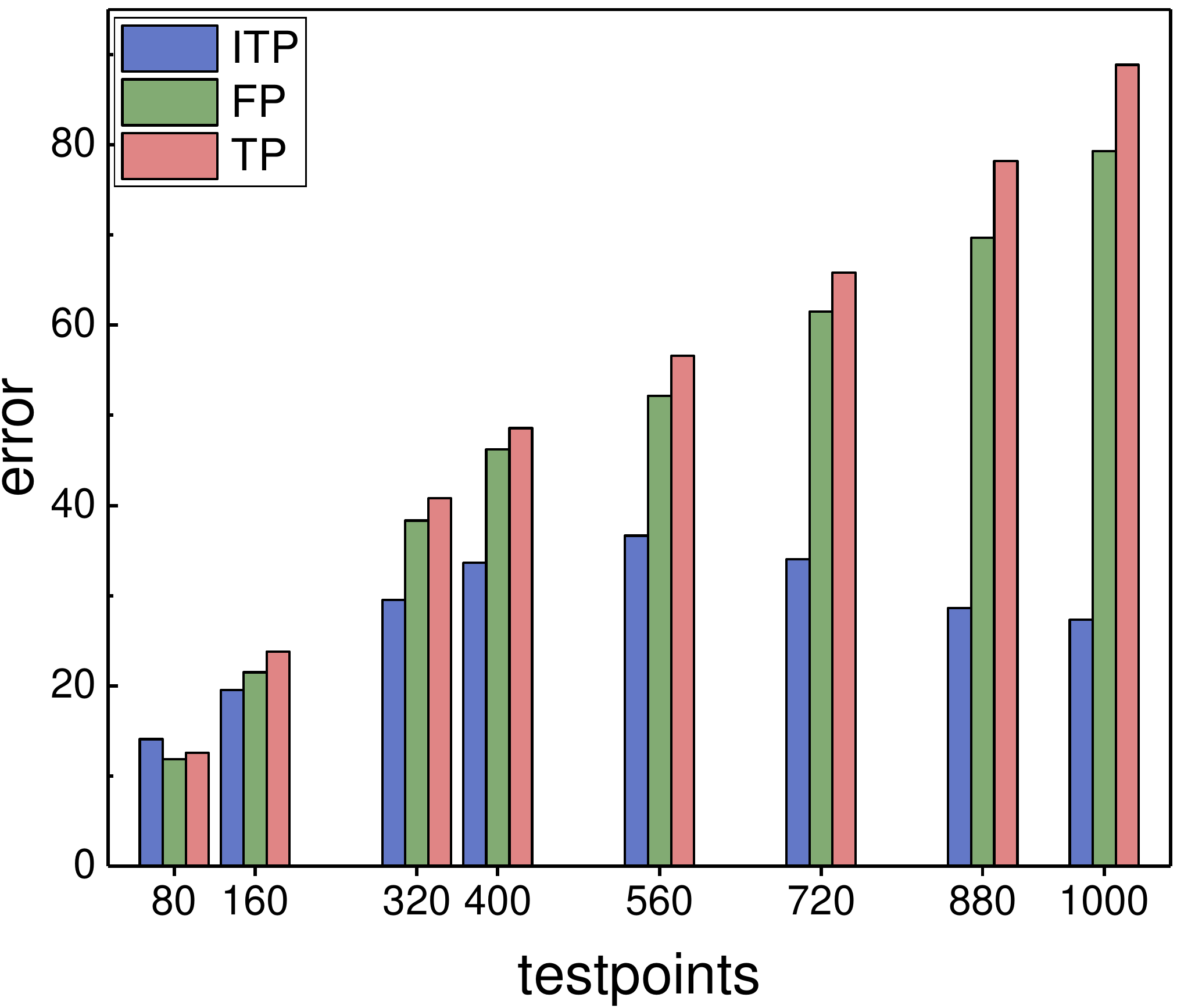}}
  \quad \quad
  \subfigure[]{
  \includegraphics[width=0.45\linewidth]{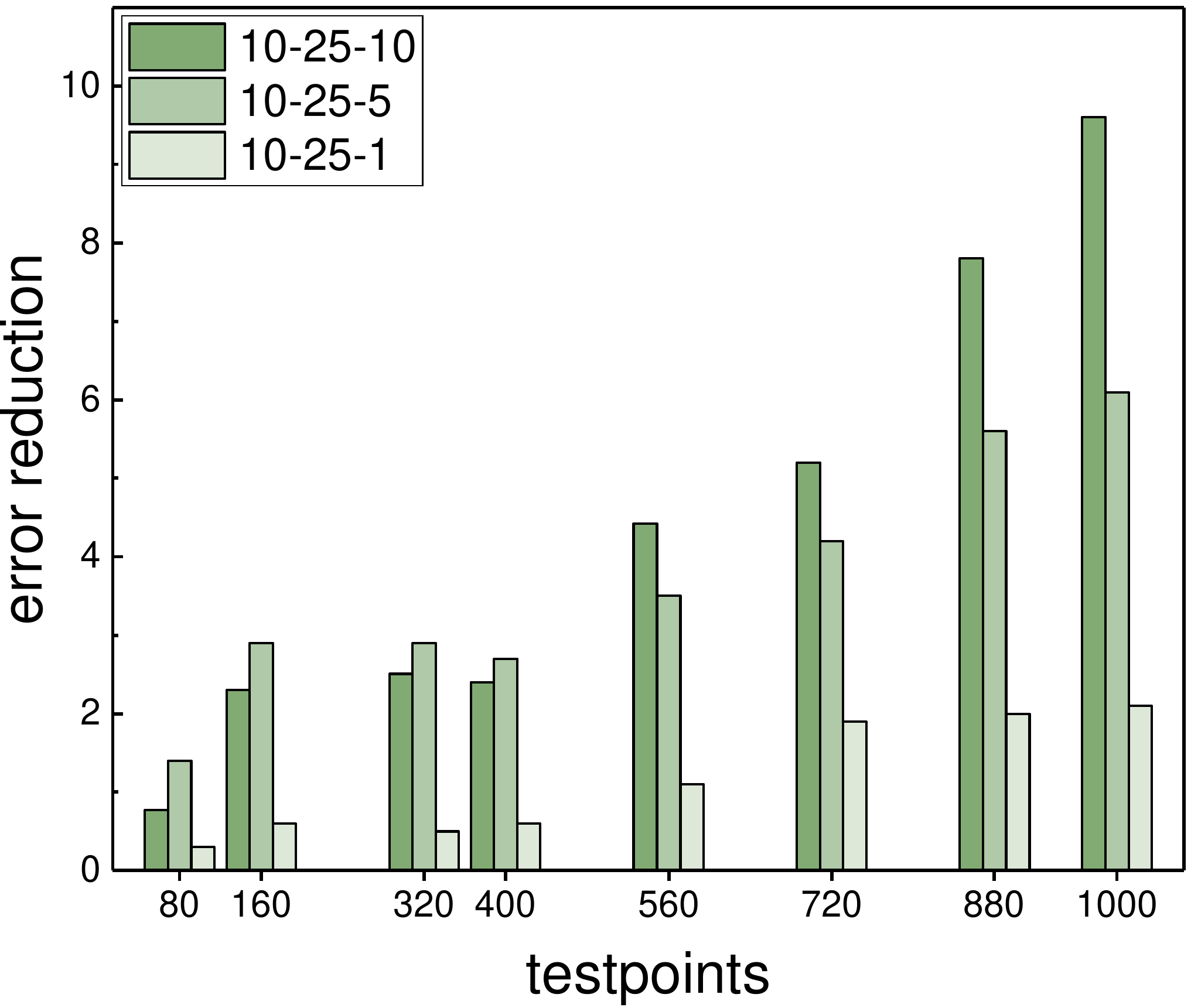}}
  \caption{(a) The average MPJPE 10-25-10 interpolated error from ITP, final predicted error from FP, and error obtained under traditional prediction pattern (TP) on CMU-Mocap. (b) Comparison of the advancement achieved with different lengths of PK on CMU-Mocap.}
  \label{ab}
\end{figure}

\noindent\textbf{Interpolation results.} During the training of ITP, the loss between the interpolated sequence and the corresponding ground truth is minimized. In Figure \ref{ab} (a), among the eight testpoints within the range of 1000ms, the interpolation error show the performance of first increasing then decreasing, which is consistent with the variation trend of general interpolation problems, and the final prediction error appears a continuous increasing trend as expected.

A phenomenon easily neglected has emerged, that the interpolation process yields higher error than extrapolation at the beginning testpoint (i.e. 80ms), which seemingly contradicts our previous statement that interpolation often yields better results than extrapolation. In fact, human motions are of highly smoothness with little difference between adjacent frames, and therefore the beginning predicted poses are quite close to observations. In other words, 10 poses of observation is sufficient to predict the following few frames directly. However, when the Priv-Encoder is involved in ITP to form an interpolation structure, the Obs-Encoder is compelled to sacrifice part of its original predictive property, to collaborate with Priv-Encoder for an \emph{overall} (especially long-term) better prediction performance and thereby a relatively accurate representation of PK. Concerning FP-Network, where we reduce the weight of PKSL, it is acceptable even if the PK simulation is not \emph{that} accurate, as PK just serves as an assistance in the extrapolation process, which leads to a continuous error-increase trend as standard extrapolation does, but lower than that without PK assisted. We regard the traditional prediction pattern that has no PK involved as TP, and the discrepancy between FP and TP is due to the distilled PK taking effects.

\noindent\textbf{Privileged sequences with different lengths.} We further set the length of the privileged sequence to 10, 5, 1 and 0, to evaluate how different lengths of privileged poses influence the final prediction performance. When the privileged length is down to 0, the two-step structure vanishes and is equivalent to predicting directly without PK (i.e. TP). In Figure \ref{ab} (b), we plot the advancement of PK-GCN trained with PK obtained from different lengths of privileged sequences on CMU-Mocap. From the figure, once PK is introduced, the final prediction performance is improved even when there only exists 1 privileged pose of PK. The higher error reduction of 10-25-5 than 10-25-10 at early testpoints indicates that too much PK may become a burden for short-term prediction.

\begin{table*}[ht]
  \caption{Comparisons of MPJPE error of 10-25-10 prediction on baselines with/without PSL and our PK-GCN on CMU-Mocap.}
  \label{ab-3}
  \centering
  \renewcommand{\arraystretch}{1.1}
  \setlength\tabcolsep{3.2pt}
  \scalebox{0.82}{
    \begin{tabular}{cc|cccccc}
      \multicolumn{2}{c|}{} & \multicolumn{6}{c}{Average} \\
      \multicolumn{2}{c|}{millisecond (ms)} & \multicolumn{1}{c|}{80} & \multicolumn{1}{c|}{160} & \multicolumn{1}{c|}{320} & \multicolumn{1}{c|}{400} & \multicolumn{1}{c|}{560} & 1000 \\ \hline
      \multicolumn{1}{c|}{\multirow{2}{*}{Traj-GCN \cite{27}}} & w/o PSL & \multicolumn{1}{c|}{13.9} & \multicolumn{1}{c|}{24.8} & \multicolumn{1}{c|}{43.3} & \multicolumn{1}{c|}{52.7} & \multicolumn{1}{c|}{55.8} & 86.2 \\ \cline{2-8} 
      \multicolumn{1}{c|}{} & 1*PSL/0.6*PSL & \multicolumn{1}{c|}{14.3/14.2} & \multicolumn{1}{c|}{26.0/25.3} & \multicolumn{1}{c|}{44.6/43.9} & \multicolumn{1}{c|}{54.2/53.7} & \multicolumn{1}{c|}{57.6/56.6} & 88.4/87.2 \\ \hline \hline
      \multicolumn{1}{c|}{\multirow{2}{*}{MSR-GCN \cite{30-2}}} & w/o PSL & \multicolumn{1}{c|}{\textbf{10.5}} & \multicolumn{1}{c|}{23.1} & \multicolumn{1}{c|}{39.7} & \multicolumn{1}{c|}{48.8} & \multicolumn{1}{c|}{53.7} & 83.0 \\ \cline{2-8} 
      \multicolumn{1}{c|}{} & 1*PSL/0.6*PSL & \multicolumn{1}{c|}{11.2/10.6} & \multicolumn{1}{c|}{24.0/23.6} & \multicolumn{1}{c|}{41.6/39.3} & \multicolumn{1}{c|}{49.5/48.8} & \multicolumn{1}{c|}{55.5/54.3} & 85.6/85.0 \\ \hline \hline
      \multicolumn{2}{c|}{PK-GCN} & \multicolumn{1}{c|}{11.8} & \multicolumn{1}{c|}{\textbf{21.5}} & \multicolumn{1}{c|}{\textbf{38.3}} & \multicolumn{1}{c|}{\textbf{46.2}} & \multicolumn{1}{c|}{\textbf{52.2}} & \textbf{79.3} \\ \hline
      \end{tabular}}
\end{table*}

\noindent\textbf{Baselines with PK.} In order to a) explore the impact of introducing PK on baselines under the traditional prediction pattern, and b) dispel the uncertainty whether our progress is \emph{only} due to PK while irrelevant to our two-step structure, we re-train Traj-GCN \cite{27} and MSR-GCN \cite{30-2} by also predicting poses in the privileged window and adding a loss (i.e. Privileged Sequence Loss, PSL) on them. Table \ref{ab-3} shows errors of baselines with/without PSL for 10-25-10 prediction. We further involve a reduction of PSL weight to 0.6, in line with the weight $ \lambda $ of PKSL in PK-GCN.

From the table, baselines with PSL yield even higher errors than without PSL, where the gap in-between exactly reflects the superiority of our two-step design. For baselines, when introducing PSL, the minimization of the total loss transforms the original 25-pose prediction into a 35-pose one. Since extrapolation becomes much harder when the prediction goes far, the last 10 poses, not belonging to the prediction object, waste big proportion of model resource and therefore lead to a worse performance of the previous 25 poses. Although an alleviation on PSL weight provides limited improvement, the direct use of PSL still harms the 25-pose prediction. However, our two-step spirit allows PK-GCN to remove these wastes during the training of FP-Network, focusing mainly on the prediction of 25 poses as we need.

\section{Conclusion}

In this paper, we introduce the overlooked poses existing after the predicted sequence to provide PK for human motion prediction from the view of interpolation. We extend the research window to the observed-predicted-privileged sequence, and propose a two-step model PK-GCN, wherein the ITP-Network learns a PK representation while interpolating the middle sequence, and the FP-Network predicts with the distilled PK assisted. Moreover, a novel PK-Simulator embedded in FP approximates the PK representation to implement the distillation process, enabling FP to imitate the behavior of ITP, and thereby keep the merits of interpolation during the final prediction. Our model outperforms the state-of-the-art methods on three benchmarked datasets, especially those activities with large ranges of motions.

~\\
\noindent\textbf{Acknowledgements.} This work was supported in part by the National Natural Science Foundation of China (NO. 62176125, 61772272).

\clearpage
%
%
\bibliographystyle{splncs04}
\bibliography{egbib}

\begin{thebibliography}{10}
\providecommand{\url}[1]{\texttt{#1}}
\providecommand{\urlprefix}{URL }
\providecommand{\doi}[1]{https://doi.org/#1}

\bibitem{52}
Aliakbarian, S., Saleh, F.S., Salzmann, M., Petersson, L., Gould, S.: A
  stochastic conditioning scheme for diverse human motion prediction. In: CVPR.
  pp. 5223--5232 (2020)

\bibitem{8-1}
Bai, S., Kolter, J.Z., Koltun, V.: An empirical evaluation of generic
  convolutional and recurrent networks for sequence modeling. arXiv preprint
  arXiv:1803.01271  (2018)

\bibitem{28}
Butepage, J., Black, M.J., Kragic, D., Kjellstrom, H.: Deep representation
  learning for human motion prediction and classification. In: CVPR. pp.
  6158--6166 (2017)

\bibitem{31-3}
Cho, J.H., Hariharan, B.: On the efficacy of knowledge distillation. In: ICCV.
  pp. 4794--4802 (2019)

\bibitem{26-1}
Corona, E., Pumarola, A., Alenyà, G., Moreno-Noguer, F.: Context-aware human
  motion prediction. In: CVPR. pp. 6992--7001 (2020)

\bibitem{5-3}
Cui, Q., Sun, H., Li, Y., Kong, Y.: A deep bi-directional attention network for
  human motion recovery. In: IJCAI. pp. 701--707 (2019)

\bibitem{29}
Cui, Q., Sun, H., Yang, F.: Learning dynamic relationships for 3d human motion
  prediction. In: CVPR. pp. 6519--6527 (2020)

\bibitem{30-2}
Dang, L., Nie, Y., Long, C., Zhang, Q., Li, G.: Msr-gcn: Multi-scale residual
  graph convolution networks for human motion prediction. In: ICCV. pp.
  11467--11476 (2021)

\bibitem{tnn}
Dong, M., Xu, C.: Skeleton-based human motion prediction with privileged
  supervision. IEEE Transactions on Neural Networks and Learning Systems
  (2022)

\bibitem{56}
Fernando, B., Herath, S.: Anticipating human actions by correlating past with
  the future with jaccard similarity measures. In: CVPR. pp. 13224--13233
  (2021)

\bibitem{5-1}
Fragkiadaki, K., Levine, S., Felsen, P., Malik, J.: Recurrent network models
  for human dynamics. In: ICCV. pp. 4346--4354 (2015)

\bibitem{18}
Gopalakrishnan, A., Mali, A., Kifer, D., Giles, L., Ororbia, A.G.: A neural
  temporal model for human motion prediction. In: CVPR. pp. 12116--12125 (2019)

\bibitem{3}
Gu, C., Sun, C., Ross, D.A., Vondrick, C., Pantofaru, C., Li, Y.,
  Vijayanarasimhan, S., Toderici, G., Ricco, S., Sukthankar, R., et~al.: Ava: A
  video dataset of spatio-temporally localized atomic visual actions. In: CVPR.
  pp. 6047--6056 (2018)

\bibitem{7-2}
Gui, L.Y., Wang, Y.X., Liang, X., Moura, J.M.: Adversarial geometry-aware human
  motion prediction. In: ECCV. pp. 786--803 (2018)

\bibitem{54}
Hernandez, A., Gall, J., Moreno-Noguer, F.: Human motion prediction via
  spatio-temporal inpainting. In: ICCV. pp. 7134--7143 (2019)

\bibitem{12}
Hinton, G., Vinyals, O., Dean, J.: Distilling the knowledge in a neural
  network. arXiv preprint arXiv:1503.02531  (2015)

\bibitem{14}
Hong, M., Xie, Y., Li, C., Qu, Y.: Distilling image dehazing with heterogeneous
  task imitation. In: CVPR. pp. 3462--3471 (2020)

\bibitem{37}
Ionescu, C., Papava, D., Olaru, V., Sminchisescu, C.: Human3.6m: Large scale
  datasets and predictive methods for 3d human sensing in natural environments.
  IEEE transactions on pattern analysis and machine intelligence
  \textbf{36}(7),  1325--1339 (2013)

\bibitem{5-2}
Jain, A., Zamir, A.R., Savarese, S., Saxena, A.: Structural-rnn: Deep learning
  on spatio-temporal graphs. In: CVPR. pp. 5308--5317 (2016)

\bibitem{49}
Kingma, D.P., Ba, J.: Adam: A method for stochastic optimization. arXiv
  preprint arXiv:1412.6980  (2014)

\bibitem{2}
Koppula, H.S., Saxena, A.: Anticipating human activities for reactive robotic
  response. In: IROS. p.~2071 (2013)

\bibitem{7-1}
Li, C., Zhang, Z., Lee, W.S., Lee, G.H.: Convolutional sequence to sequence
  model for human dynamics. In: CVPR. pp. 5226--5234 (2018)

\bibitem{30-1}
Li, M., Chen, S., Zhao, Y., Zhang, Y., Wang, Y., Tian, Q.: Dynamic multiscale
  graph neural networks for 3d skeleton based human motion prediction. In:
  CVPR. pp. 214--223 (2020)

\bibitem{26-2}
Liang, M., Yang, B., Hu, R., Chen, Y., Liao, R., Feng, S., Urtasun, R.:
  Learning lane graph representations for motion forecasting. In: ECCV. pp.
  541--556 (2020)

\bibitem{22}
Liu, J., Shahroudy, A., Xu, D., Wang, G.: Spatio-temporal lstm with trust gates
  for 3d human action recognition. In: ECCV. pp. 816--833 (2016)

\bibitem{trajectorycnn}
Liu, X., Yin, J., Liu, J., Ding, P., Liu, J., Liu, H.: Trajectorycnn: a new
  spatio-temporal feature learning network for human motion prediction. IEEE
  Transactions on Circuits and Systems for Video Technology  \textbf{31}(6),
  2133--2146 (2020)

\bibitem{34}
Liu, Y., Chen, K., Liu, C., Qin, Z., Luo, Z., Wang, J.: Structured knowledge
  distillation for semantic segmentation. In: CVPR. pp. 2604--2613 (2019)

\bibitem{cues}
Liu, Z., Su, P., Wu, S., Shen, X., Chen, H., Hao, Y., Wang, M.: Motion
  prediction using trajectory cues. In: ICCV. pp. 13299--13308 (2021)

\bibitem{KD-hmp}
Ma, H., Li, J., Hosseini, R., Tomizuka, M., Choi, C.: Multi-objective diverse
  human motion prediction with knowledge distillation. In: CVPR. pp. 8161--8171
  (2022)

\bibitem{progressively}
Ma, T., Nie, Y., Long, C., Zhang, Q., Li, G.: Progressively generating better
  initial guesses towards next stages for high-quality human motion prediction.
  In: CVPR. pp. 6437--6446 (2022)

\bibitem{41}
Mao, W., Liu, M., Salzmann, M.: History repeats itself: Human motion prediction
  via motion attention. In: ECCV. pp. 474--489 (2020)

\bibitem{53}
Mao, W., Liu, M., Salzmann, M.: Generating smooth pose sequences for diverse
  human motion prediction. In: ICCV. pp. 13309--13318 (2021)

\bibitem{mao2022}
Mao, W., Liu, M., Salzmann, M.: Weakly-supervised action transition learning
  for stochastic human motion prediction. In: CVPR. pp. 8151--8160 (2022)

\bibitem{27}
Mao, W., Liu, M., Salzmann, M., Li, H.: Learning trajectory dependencies for
  human motion prediction. In: ICCV. pp. 9489--9497 (2019)

\bibitem{21}
Martinez, J., Black, M.J., Romero, J.: On human motion prediction using
  recurrent neural networks. In: CVPR. pp. 2891--2900 (2017)

\bibitem{13}
Mishra, A., Marr, D.: Apprentice: Using knowledge distillation techniques to
  improve low-precision network accuracy. arXiv preprint arXiv:1711.05852
  (2017)

\bibitem{KD-tp}
Monti, A., Porrello, A., Calderara, S., Coscia, P., Ballan, L., Cucchiara, R.:
  How many observations are enough? knowledge distillation for trajectory
  forecasting. In: CVPR. pp. 6553--6562 (2022)

\bibitem{1}
Paden, B., {\v{C}}{\'a}p, M., Yong, S.Z., Yershov, D., Frazzoli, E.: A survey
  of motion planning and control techniques for self-driving urban vehicles.
  IEEE Transactions on intelligent vehicles  \textbf{1}(1),  33--55 (2016)

\bibitem{48}
Paszke, A., Gross, S., Chintala, S., Chanan, G., Yang, E., DeVito, Z., Lin, Z.,
  Desmaison, A., Antiga, L., Lerer, A.: Automatic differentiation in pytorch
  (2017)

\bibitem{quaternion}
Pavllo, D., Feichtenhofer, C., Auli, M., Grangier, D.: Modeling human motion
  with quaternion-based neural networks. International Journal of Computer
  Vision  \textbf{128}(4),  855--872 (2020)

\bibitem{33}
Romero, A., Ballas, N., Kahou, S.E., Chassang, A., Gatta, C., Bengio, Y.:
  Fitnets: Hints for thin deep nets. arXiv preprint arXiv:1412.6550  (2014)

\bibitem{32-2}
Shen, Z., He, Z., Xue, X.: Meal: Multi-model ensemble via adversarial learning.
  In: AAAI. pp. 4886--4893 (2019)

\bibitem{60}
Sofianos, T., Sampieri, A., Franco, L., Galasso, F.: Space-time-separable graph
  convolutional network for pose forecasting. In: ICCV. pp. 11209--11218 (2021)

\bibitem{guided}
Sun, J., Lin, Z., Han, X., Hu, J.F., Xu, J., Zheng, W.S.: Action-guided 3d
  human motion prediction. NeurIPS  \textbf{34},  30169--30180 (2021)

\bibitem{32-1}
Tung, F., Mori, G.: Similarity-preserving knowledge distillation. In: ICCV. pp.
  1365--1374 (2019)

\bibitem{57}
Vapnik, V., Vashist, A.: A new learning paradigm: Learning using privileged
  information. Neural Networks  \textbf{22}(5-6),  544--557 (2009)

\bibitem{3dpw}
Von~Marcard, T., Henschel, R., Black, M.J., Rosenhahn, B., Pons-Moll, G.:
  Recovering accurate 3d human pose in the wild using imus and a moving camera.
  In: ECCV. pp. 601--617 (2018)

\bibitem{16}
Wang, T., Yuan, L., Zhang, X., Feng, J.: Distilling object detectors with
  fine-grained feature imitation. In: CVPR. pp. 4933--4942 (2019)

\bibitem{58}
Xu, M., Gao, M., Chen, Y.T., Davis, L.S., Crandall, D.J.: Temporal recurrent
  networks for online action detection. In: CVPR. pp. 5532--5541 (2019)

\bibitem{8-2}
Yan, S., Xiong, Y., Lin, D.: Spatial temporal graph convolutional networks for
  skeleton-based action recognition. In: AAAI (2018)

\bibitem{31-4}
Yang, C., Xie, L., Su, C., Yuille, A.L.: Snapshot distillation: Teacher-student
  optimization in one generation. In: CVPR. pp. 2859--2868 (2019)

\bibitem{59}
Yuan, Y., Kitani, K.: Dlow: Diversifying latent flows for diverse human motion
  prediction. In: ECCV. pp. 346--364 (2020)

\bibitem{11}
Zhao, P., Xie, L., Zhang, Y., Wang, Y., Tian, Q.: Privileged knowledge
  distillation for online action detection. arXiv preprint arXiv:2011.09158
  (2020)

\bibitem{gating}
Zhong, C., Hu, L., Zhang, Z., Ye, Y., Xia, S.: Spatio-temporal gating-adjacency
  gcn for human motion prediction. In: CVPR. pp. 6447--6456 (2022)

\end{thebibliography}
\end{document}